# The Open Corpus of the Veps and Karelian Languages: Overview and Applications

**Tatyana Boyko[1], Nina Zaitseva[1], Natalia Krizhanovskaya[2], Andrew Krizhanovsky[2,3,*], Irina Novak[1], Nataliya Pellinen[1], and Aleksandra Rodionova[1]**

[1] Institute of Linguistics, Literature and History, Karelian Research Centre, Russian Academy of Sciences
[2] Institute of Applied Mathematical Research, Karelian Research Centre, Russian Academy of Sciences
[3] Petrozavodsk State University
[*]Corresponding author: Andrew Krizhanovsky; Email: andrew.krizhanovsky@gmail.com; ORCID: 0000-0003-3717-2079

Other ORCID iDs: Tatyana Boyko ORCID: 0000-0001-5095-2921; Nina Zaitseva ORCID: 0000-0002-8335-2137; Natalia Krizhanovskaya ORCID: 0000-0002-9948-1910; Irina Novak ORCID: 0000-0002-9436-9460; Nataliya Pellinen ORCID: 0000-0002-5648-6877; Aleksandra Rodionova ORCID: 0000-0001-5645-9441

**Abstract**
A growing priority in the study of Baltic-Finnic languages of the Republic of Karelia has been the methods and tools of corpus linguistics. Since 2016, linguists, mathematicians, and programmers at the Karelian Research Centre have been working with the Open Corpus of the Veps and Karelian Languages (VepKar), which is an extension of the Veps Corpus created in 2009. The VepKar corpus comprises texts in Karelian and Veps, multifunctional dictionaries linked to them, and software with an advanced system of search using various criteria of the texts (language, genre, etc.) and numerous linguistic categories (lexical and grammatical search in texts was implemented thanks to the generator of word forms that we created earlier). A corpus of 3000 texts was compiled, texts were uploaded and marked up, the system for classifying texts into languages, dialects, types and genres was introduced, and the word-form generator was created. Future plans include developing a speech module for working with audio recordings and a syntactic tagging module using morphological analysis outputs. Owing to continuous functional advancements in the corpus manager and ongoing VepKar enrichment with new material and text markup, users can handle a wide range of scientific and applied tasks. In creating the universal national VepKar corpus, its developers and managers strive to preserve and exhibit as fully as possible the state of the Veps and Karelian languages in the 19th-21st centuries.

**Keywords:** corpus linguistics, Veps language, Karelian language, national corpus, dictionary, tagging

## 1. Introduction

One of the key focuses for Baltic-Finnic linguistic studies in Karelia in the past decade has been corpus-based research on languages of the region. Although the corpus is an essential modern tool for language studies, it must be said that large annotated open-access online corpora are not always available even for languages spoken by large populations (e.g., a large Ukrainian corpus incorporating all genres appeared in 2020 [1]).

The development of corpus linguistics activities at the Institute of Linguistics, Literature and History of the Karelian Research Centre of the Russian Academy of Sciences (KarRC RAS) has been enabled by close and productive collaboration with mathematicians and programmers from the Institute of Applied Mathematical Research KarRC RAS. This effort commenced in 2009, when the Veps Corpus was created under the leadership of Nina Zaitseva[1]. The Veps Corpus consisted of five text corpora, an electronic dictionary, and search tools [2]. Digitalization of scientific knowledge, which is meant to augment data sampling, classification, and analysis possibilities, has dictated the

1 See http://vepsian.krc.karelia.ru/about/

need to widen corpus research to cover the language of the region's titular ethnic group – Karelian, and triggered the creation of the multilingual corpus since 2016. The extended corpus was entitled "The Open Corpus of the Veps and Karelian Languages" (VepKar).

During these five years, much has been done to fill the digital resource with texts as well as to supply it with dictionaries and diverse tools for conveniently using the corpus. The corpus manager in the VepKar project is the Dictorpus open source code software complex written in the PHP programming language on Laravel platform. The data are stored in a MySQL database. Significant augmentation of the VepKar corpus volume and functions is critical for linguists, who need a fully-featured research tool.

## 2. Corpus architecture and quantitative characteristics

The VepKar architecture with the latest statistics is shown in Fig. 1.

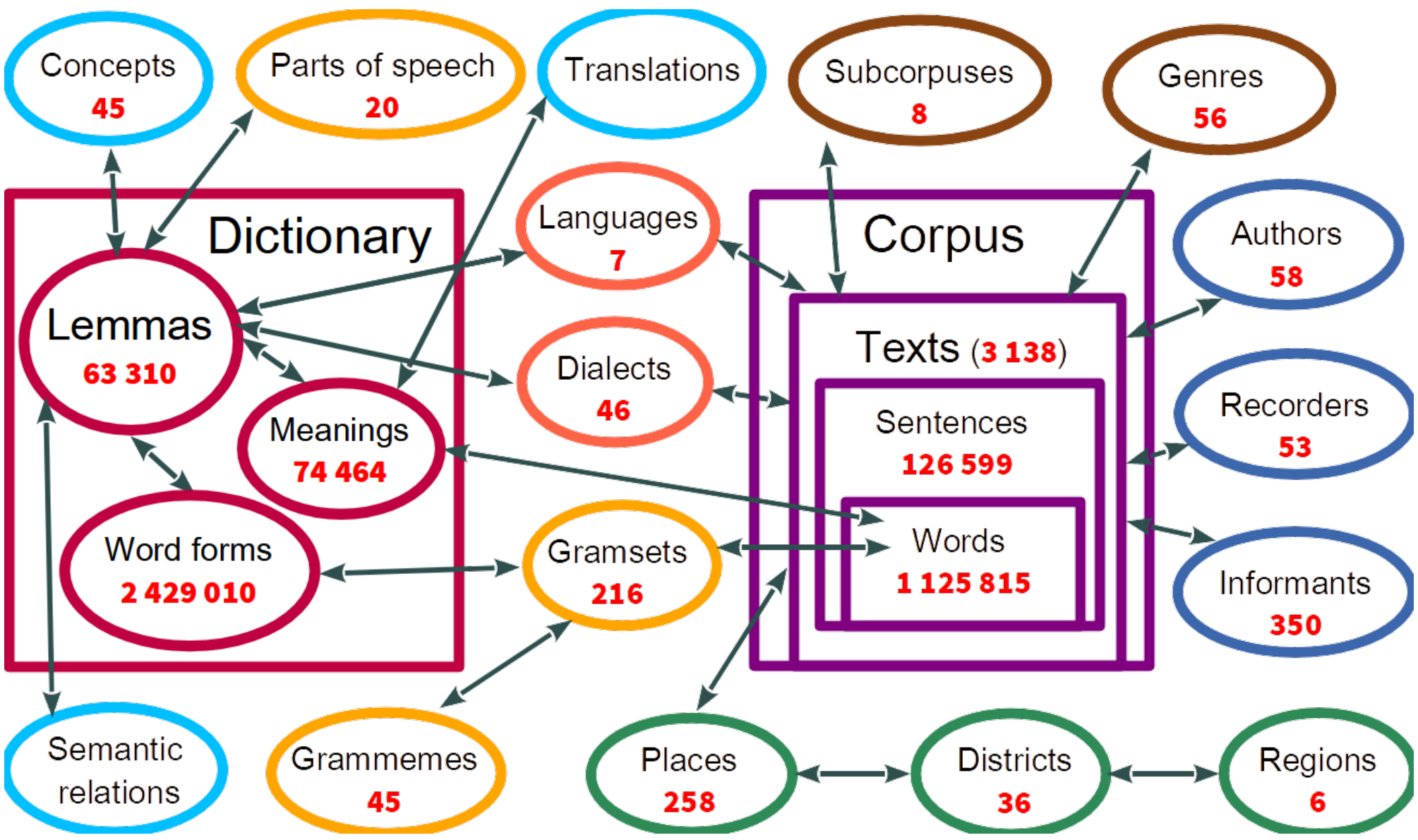

**Figure 1: Corpus architecture and quantitative characteristics**

### 2.1. Subcorpora

Expectedly, the central block of the resource is the corpus (assemblage) of texts. As of October 2021, it contains over 3000 texts, and the process of adding new texts is continuous. The list of sources for the corpus includes: published Karelian and Veps dialectal speech samples, folklore, literary, and translated texts in Karelian and Veps, material from the Karelian newspaper Oma Mua, Veps newspaper Kodima, Taival Almanac in Karelian and Verez tullei Almanac in Veps, etc. All accessions are strictly subject to approval by publishers and authors (only open-license texts are used[2]). One more apparently inexhaustible source is the KarRC RAS Scientific Archives. Continuous supplementation of the corpus with new material contributes to a wider recognition and popularity of the Karelian and Veps languages, and to successful fulfillment of various publicity, educational, and research tasks.

The process of uploading new texts to the corpus consists of several key stages:

i. selection of texts to be included in the corpus and getting permission from the author (if applicable);
ii. text digitalization (this stage is skipped for digital resources);
iii. processing and proofreading of the digitalized text to correct text recognition errors;
iv. uploading the text to the corpus (includes code conversion and segmentation processes);

2 See permissions from Veps and Karelian writers and poets for having their texts included in the VepKar corpus at http://dictorpus.krc.karelia.ru/ru/page/permission

v. metatextual markup (done manually according to template);
vi. automatic structural markup (parsing into sentences and words);
vii. automatic lexico-grammatical markup: semantic (meaning or variant of meaning) and grammatical (part of speech and the set of possible grammemes) characteristics are assigned to each token (lexical unit);
viii. revision of markup results by experts (error correction, elimination of homonymy, tagging of unrecognized tokens);
ix. parallel translation into Russian (sentence-wise).

An important element for a researcher working with a text is its metatextual markup, which includes information about the language and dialect affiliations, the author or performer (informant), dates, genre features, etc.

All the uploaded texts are distributed among subcorpora (Figure 2) according to two parameters: language affiliation and type of the text.

As the corpus combines two Baltic-Finnic languages, it is expedient to make a division into the respective subcorpora for the languages. However, the dialectal disunity of the Karelian language, which has three notably distinct supradialects, urged the developers to create three rather than one Karelian subcorpora [3]. It is within these lower-level subcorpora that texts are divided according to their dialectal affiliation (4 Veps language dialects, 25 dialects of Karelian Proper, 8 Livvi dialects, and 4 Ludian dialects) or belonging to a standardized variety of the language (one variety for Veps and four for Karelian). Standardized varieties act as a kind of a reference baseline for building the principal components of the corpus manager, permitting linguists to search through dialectal texts as well.

Speaking of the text type and style, the texts are distributed into the following major subcorpora: biblical texts, law, journalistic texts, subtitles, folklore texts, literary texts (Figure 2). Texts in the literary and folklore text subcorpora are also subdivided into genres (Figure 3).

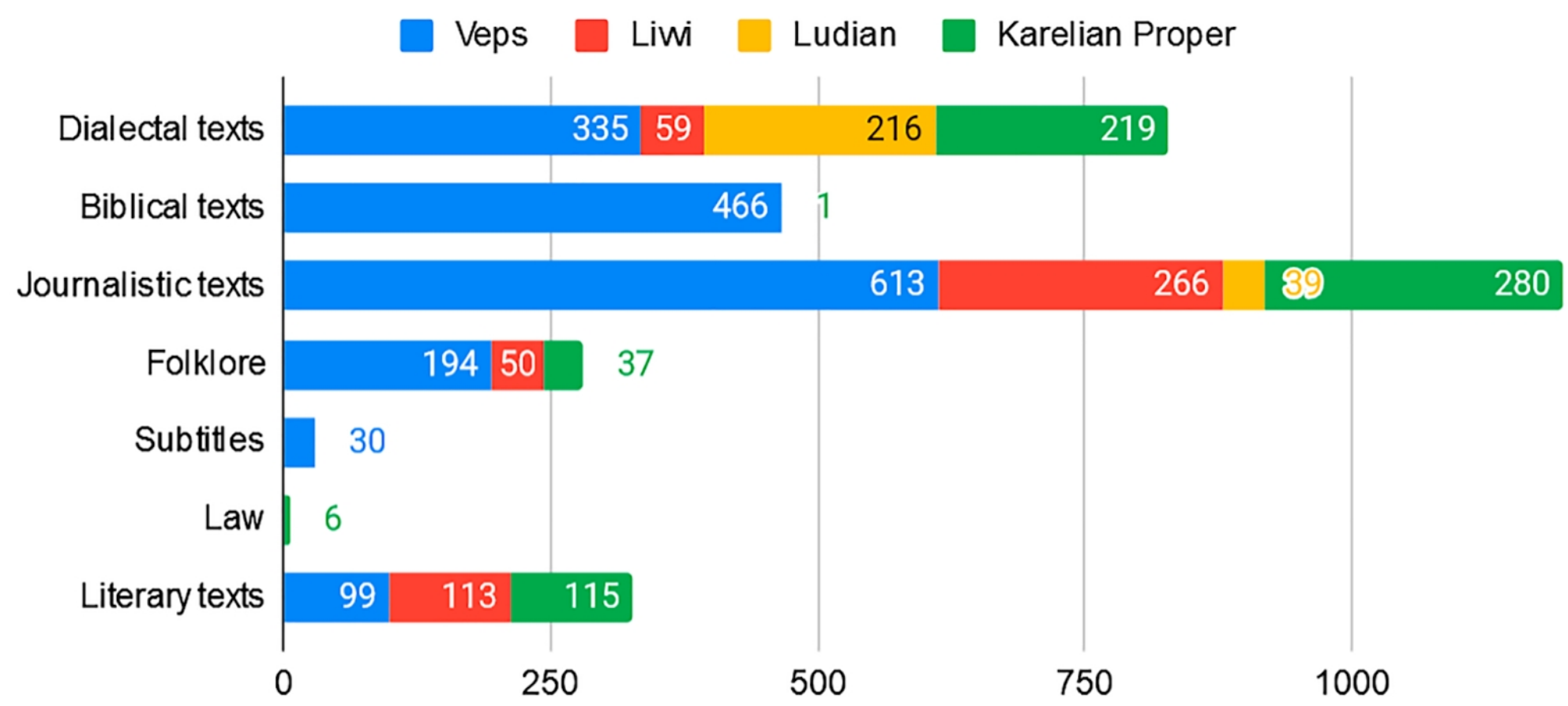

**Figure 2: Distribution of VepKar texts among subcorpora according to languages and types, as of October2021[3]**

3 For updated statistics visit http://dictorpus.krc.karelia.ru/en/stats/by_corpus

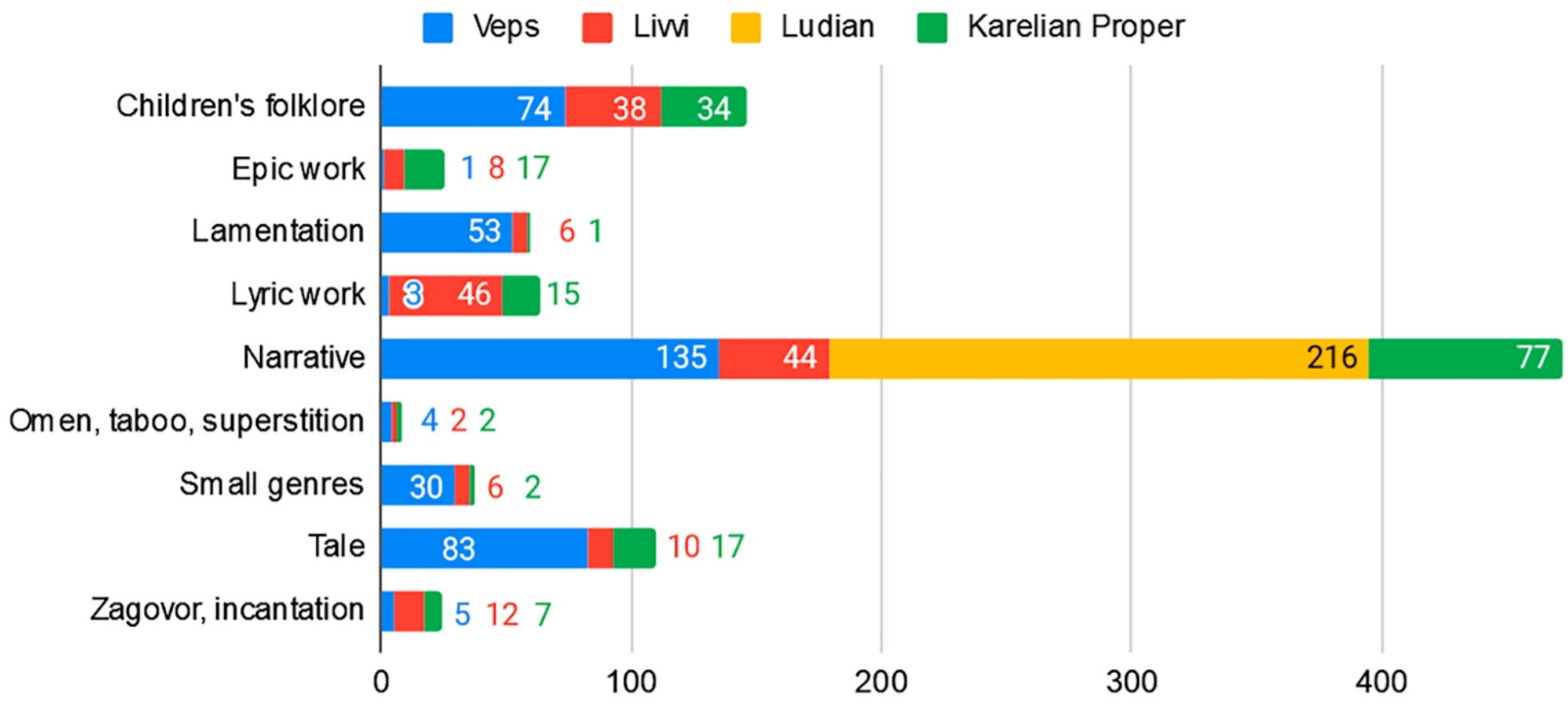

**Figure 3: Distribution of VepKar texts among genres and languages, as of October 2021**[4]

Apart from the texts, VepKar corpus has various tools for processing them. Using advanced search (Figure 4), one can filter the texts by language, type or genre as well as by informant, recorder or author, year of recording or year of publication (Figure 5).

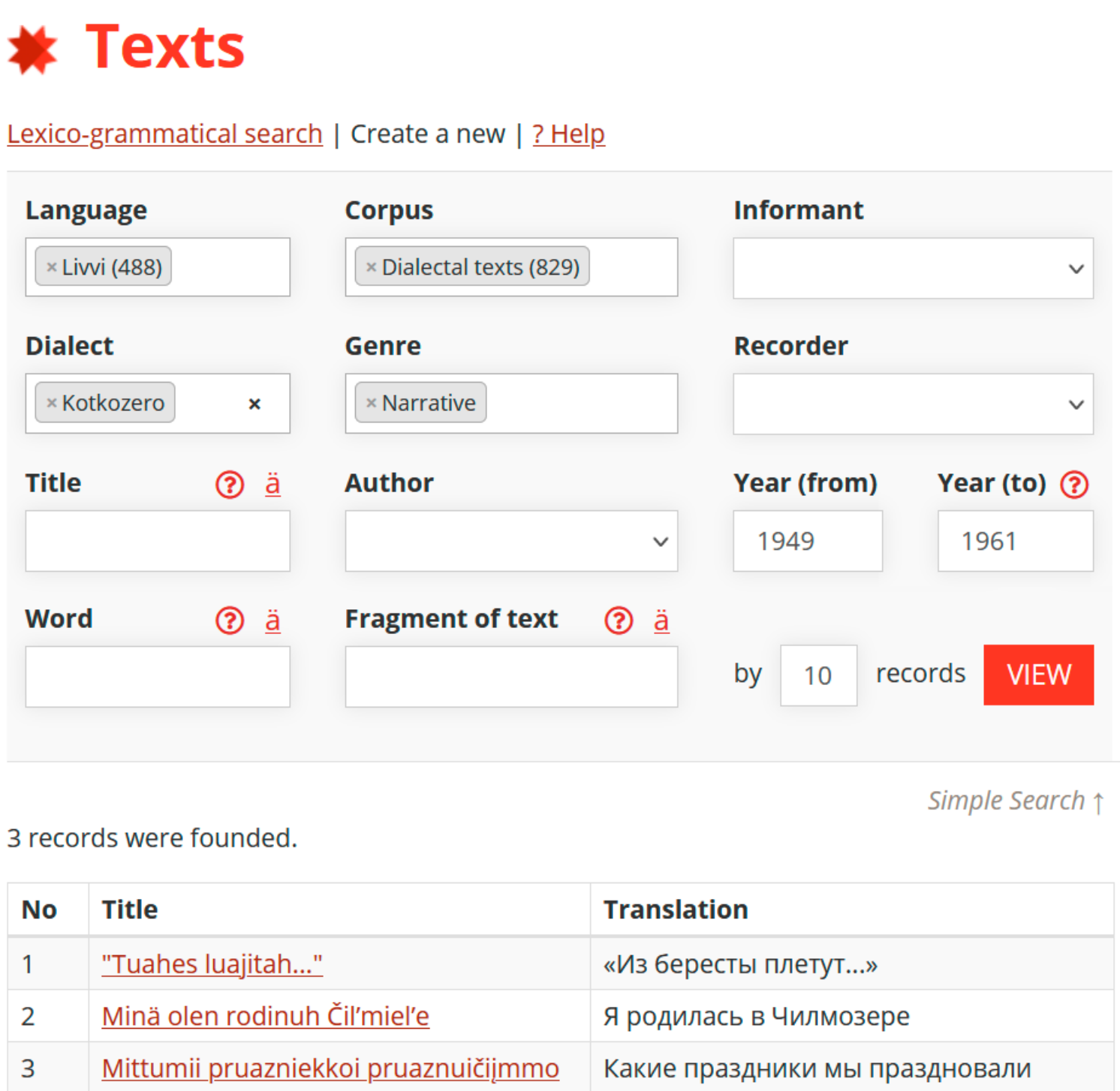

**Figure 4: Advanced search interface for VepKar texts**

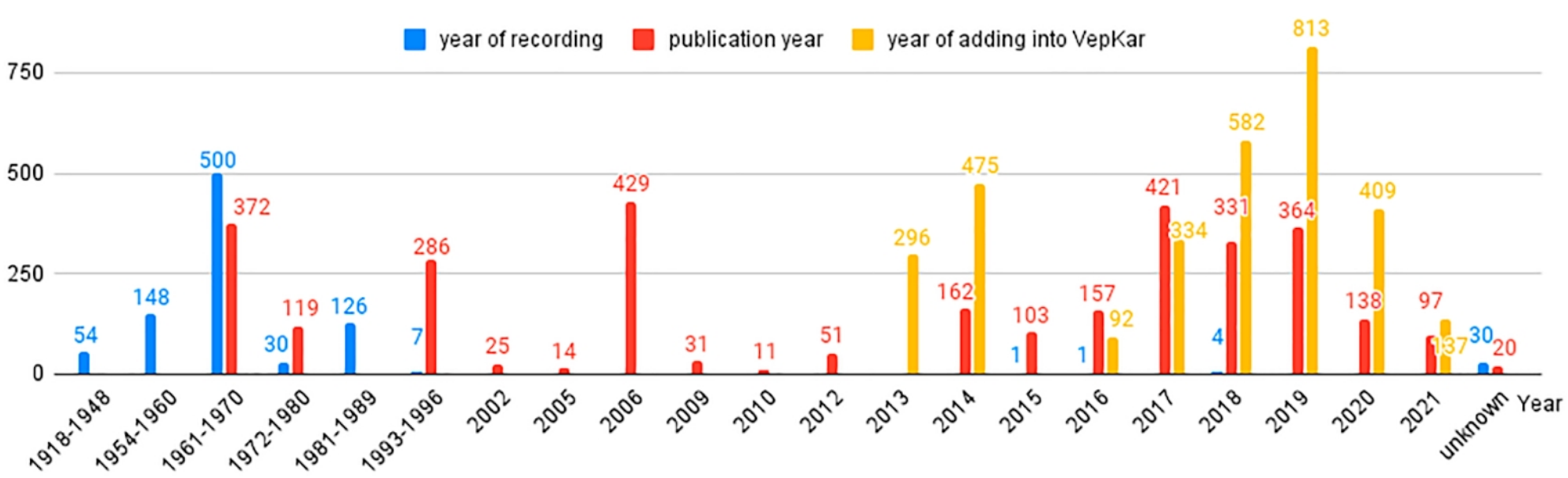

4 For updated statistics visit http://dictorpus.krc.karelia.ru/en/stats/by_genre

**Figure 5: Number of VepKar texts by year of recording, date of publication, and date of accession to the corpus, as of October 2021**[5]

### 2.2. Dictionary

The second major block that facilitates the work with the corpus of texts is the dictionary. The corpus dictionary comprises over 60 000 entries and approx. 2.5 million word forms. Words in the dictionary are mainly explained in Russian and English, but explanations can be added also in Veps, supradialects of Karelian, and in Finnish (Figure 6). The search for lemmas in the dictionary can be arranged by language and dialect affiliations, parts of speech, grammatical features, or even lexical-semantic categories (Figure 7).

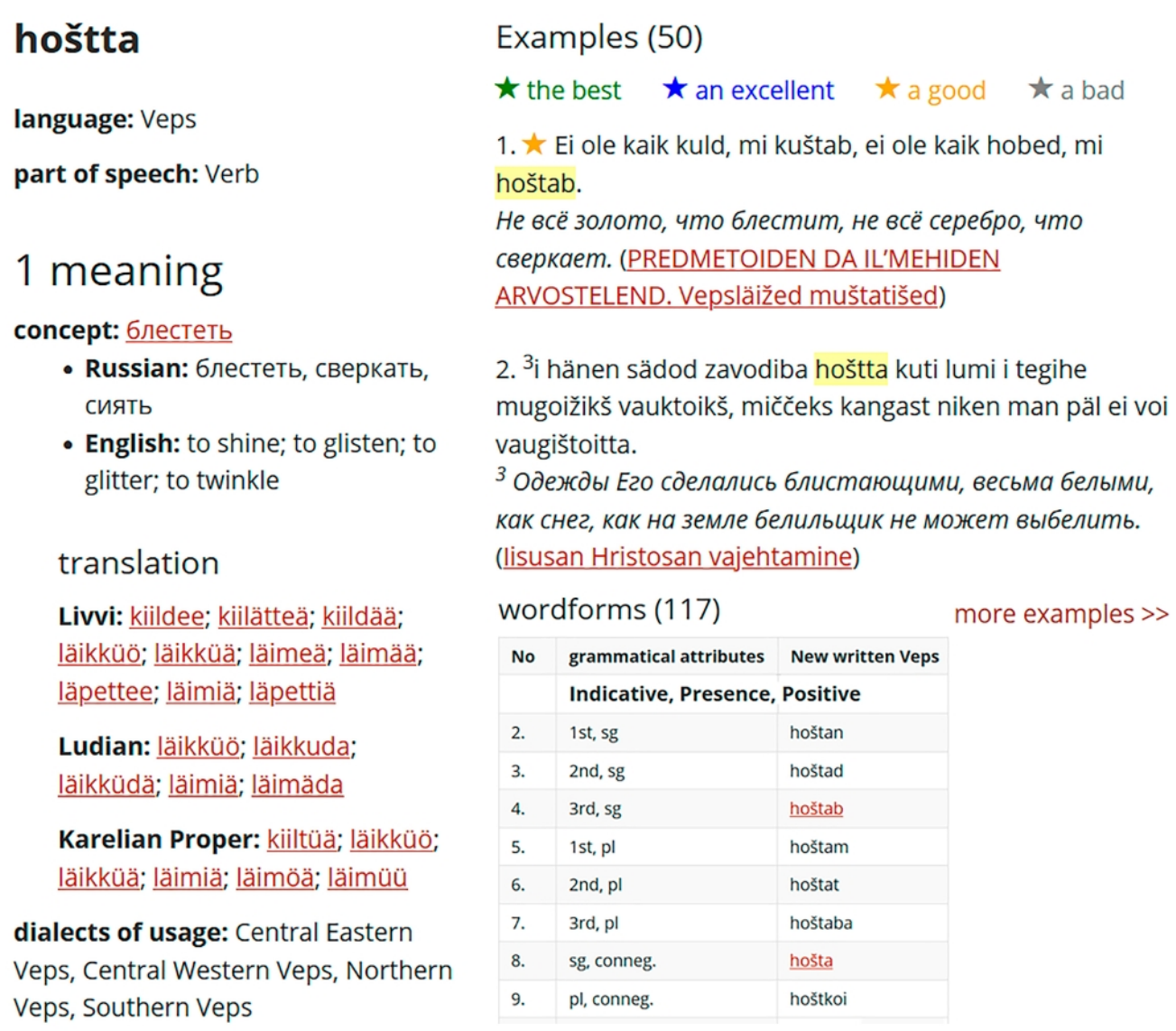

**Figure 6: Fragment of the dictionary entry hoštta with translations, examples, and word forms**[6]

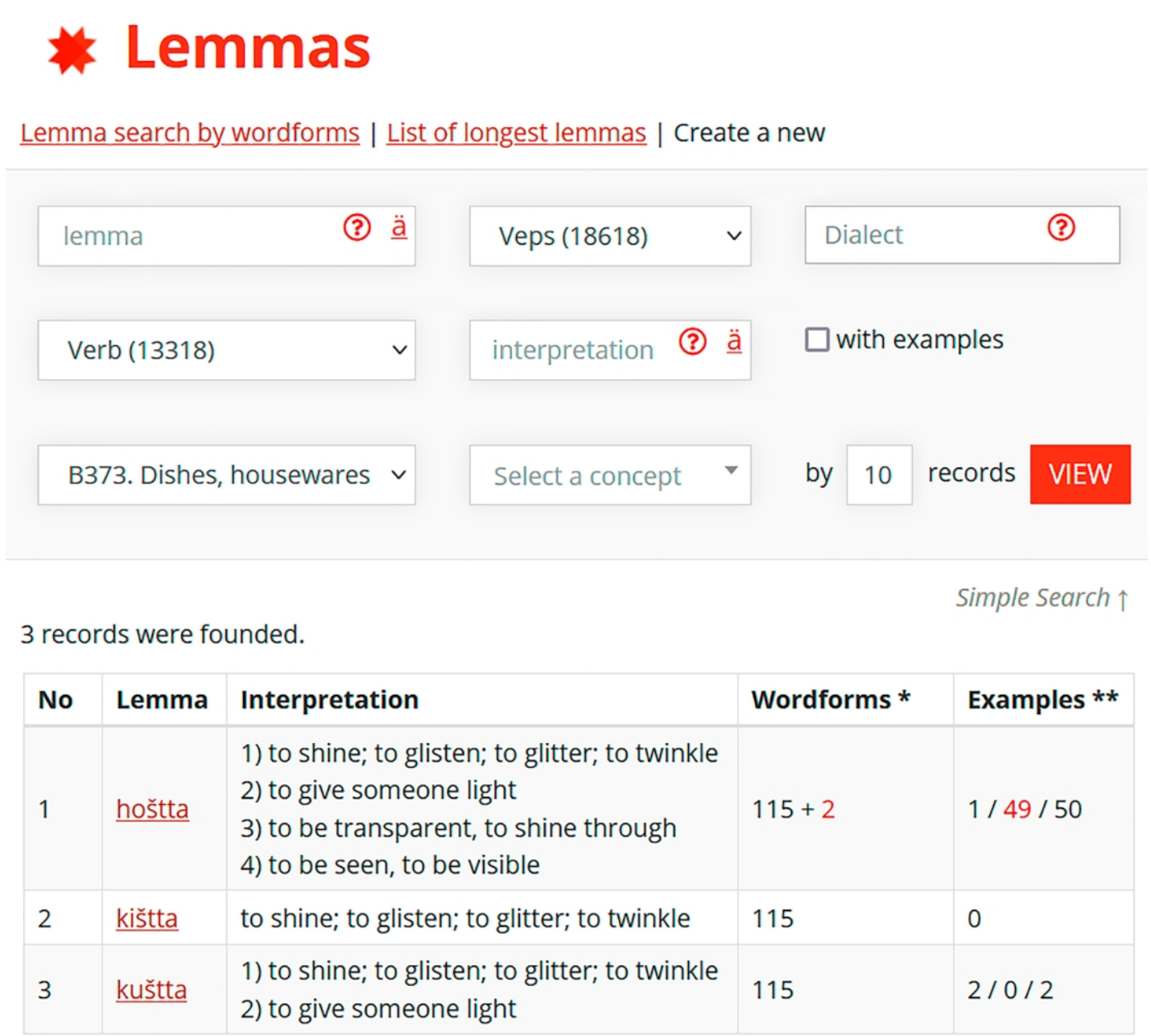

**Figure 7: Interface for the search for lemmas and the search results for Livvi verbs related to the concept of “Dishes, household utensils” in the VepKar dictionary**

5 For updated statistics visit http://dictorpus.krc.karelia.ru/en/stats/by_year
6 Full version of the entry is available at http://dictorpus.krc.karelia.ru/en/dict/lemma/1274

Lemmas and word forms with grammatical feature sets were exported from the Veps and Karelian dictionaries of the VepKar corpus to the Universal Morphological Database UniMorph 3.0. These data are available on GitHub website as separate subprojects of the project [4].

Due to the electronic format of the material, which permits retrieving various subsets and running advanced search queries, the VepKar corpus dictionary is multifunctional. It is a full-fledged interlingual dictionary (Karelian/Veps - Russian), amply illustrated due to links between lemmas (dictionary entries) and corpus texts. Owing to the search-by-explanation option, it can to a certain extent be used as an alternative to a printed Russian-Karelian or Russian-Veps dictionary. The availability of full inflectional paradigms for nominals (ca. 30 word forms) and verbs (ca. 130 word forms) gives the dictionary a grammatical-spelling perspective. In addition, full frequency- and reverse dictionaries are also available to the users. The phraseological dictionary is getting gradually filled up. The possibility of creating government (verbal, adpositional, adadverbial, adnominal, etc.) dictionaries is being considered.

Apart from its application within the corpus, VepKar dictionary is an ample source of material for all sorts of electronic Karelian and Veps dictionaries to be created. In 2021 e.g., an open source code mobile application Sanahelmi v.1 written in Kotlin language[7] was produced for Android devices[8]. This app is a simplified version of the VepKar corpus dictionary, containing all Veps and Karelian words (lemmas and word forms), grammatical features, and explanations in Russian from the VepKar dictionary.

### 2.3. Corpus-dictionary interaction

Linguistic research is facilitated by special modules of the corpus manager which connect the corpus and the dictionary. An essential feature of the VepKar corpus is automatic linguistic markup, which includes tokenization (breaking up into words), lemmatization (returning word forms to the dictionary form), and morphological markup (part-of-speech and grammatical category (gramset) identification).

The percentage of automatic markup of texts in the corpus is now 73%. This level was achieved owing to the implementation in 2019–2020 of nominal and verbal word-form generation algorithms for the Veps language, Livvi Karelian and, especially, Karelian Proper [5]. Project experts revise the markup and eliminate semantic (selection of meaning) and morphological (selection of grammatical features) homonymy (Figure 8). At the same time, a module for automatic elimination of homonymy is being developed.

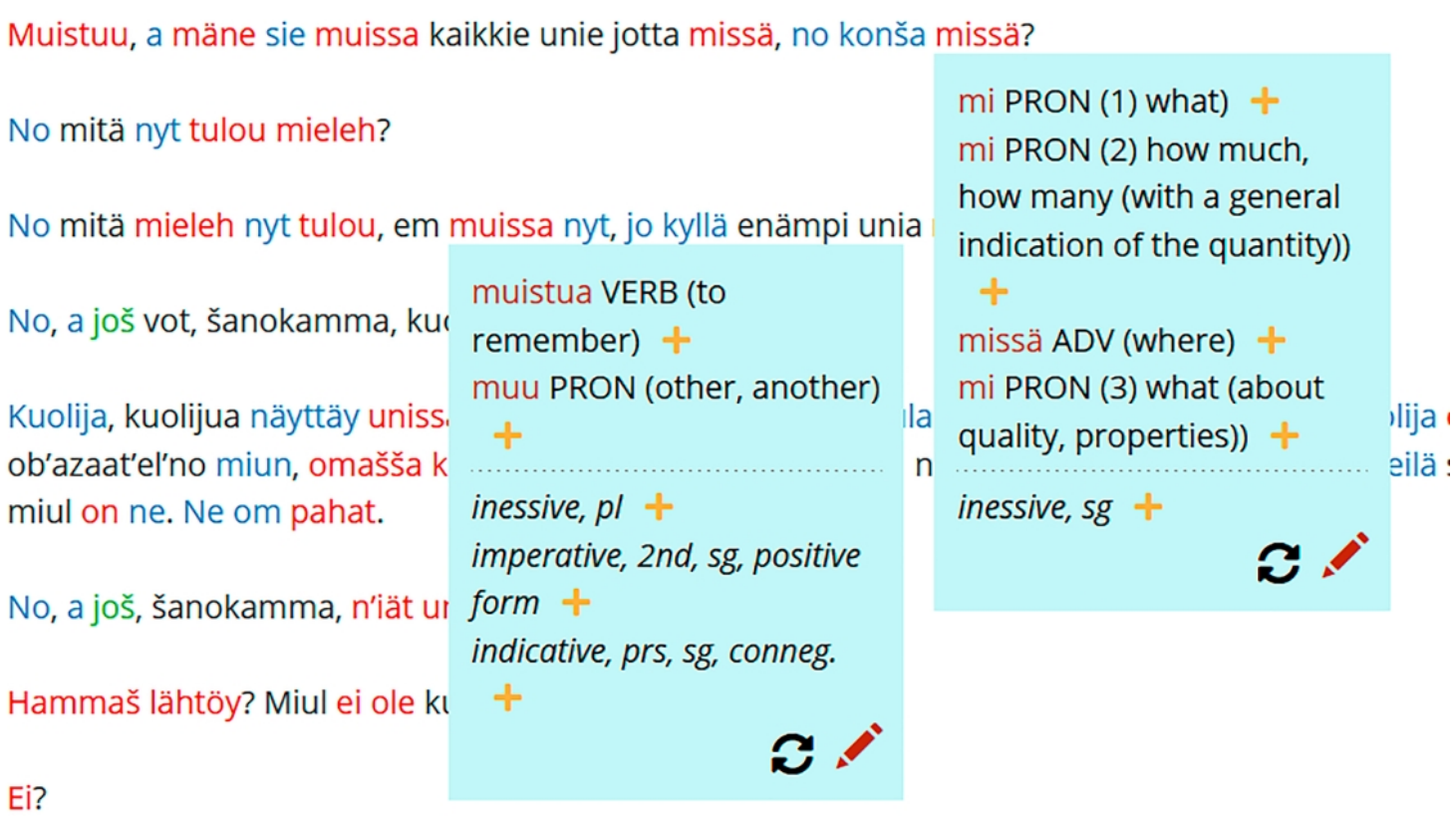

**Figure 8: Examples of semantic and morphological homonymy in the process of marking up a text in the VepKar corpus; editor chooses one of the meanings of the word and morphological features**

Interactions between the corpus and the dictionaries are facilitated by quite a number of tools. E.g., frequency dictionaries, which retrieve the most common lexemes and word forms, help editors manage the priorities in their work i.e., enter the most common lemmas into the dictionaries, thereby accelerating the rate of text markup. In addition, these dictionaries can be used for statistical research.

7 See https://github.com/componavt/sanahelmi

8 See https://play.google.com/store/apps/details?id=vepkar.test

E.g., the timeframe of the corpus permits tracking changes in the lexical composition of the Veps and Karelian languages over a century. With stylistic subcorpora available, a similar study can be implemented for certain genres or for specific texts.

In 2021, after word form generators had been created, it became possible to deliver a tool for advanced lexico-grammatical text search (Figure 9), which is especially important for dealing with morphological homonymy elimination. The predictor module gives the editor suggestions to choose from regarding parts of speech, grammatical features, and the root form of the word not recognized by automatic markup [6].

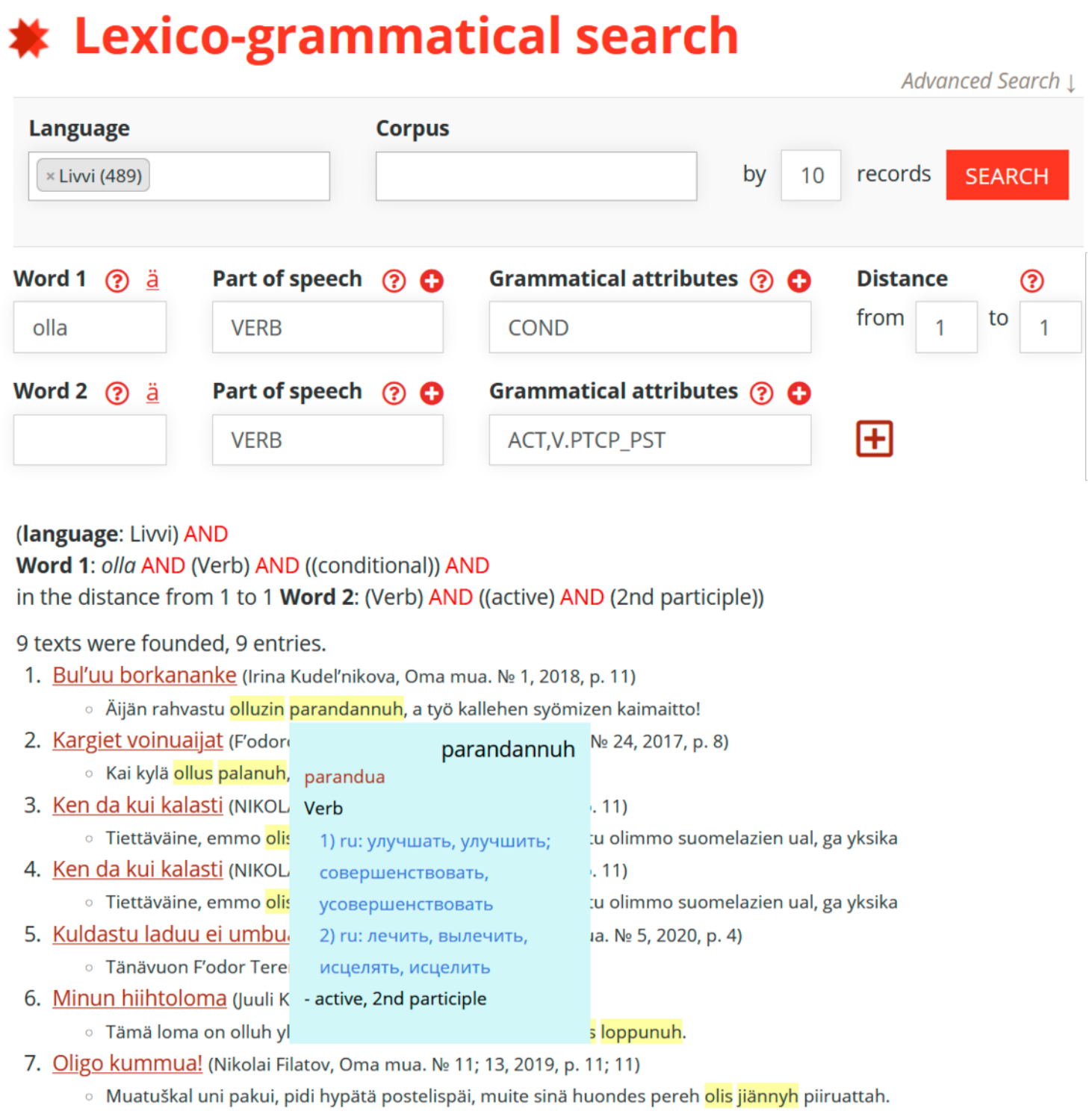

**Figure 9: Lexico-grammatical search**

A major promising task for the future is to build a morphological analyzer, which is designed e.g., to identify the endings of nominal and verbal word forms, to predict the proper forms of nominals coming with verbs (verbal government), and word order in the sentence.

Thus, corpus-dictionary interconnections and the constantly expanding functionality of the corpus manager enable multiple uses of the textual material for dealing with a variety of linguistic tasks.

## 3. Conclusions

The conclusion following from the above is that VepKar:

i. provides users with free access to full-text documents;

ii. is continuously supplemented with new data;

iii. is a multilingual resource: includes texts in Veps and Karelian (dictionaries offer explanations in Russian and partially in English);

iv. is a full-text corpus i.e., texts are fully tagged and the entire body of texts is searchable;

v. covers the languages as much as possible, i.e., includes texts of various types, genres and styles. In this sense, according to the definition suggested in [7], VepKar is the Veps and Karelian National Corpus. Furthermore, the corpus content exhibits the state of the languages in different time periods;

vi. includes the following type of markup: metatextual (text title, date of creation, author, genre, place of recording, etc.), morphological (parts of speech and morphological features are given for words in the texts), and semantic (words in texts are linked to meanings in dictionary entries).

Thus, VepKar is an open, dynamical, multilingual, written, full-text, representative national Veps and Karelian corpus containing metatextual, morphological, and semantic markup. The system has tools permitting linguists to import and export data in different formats, test and verify them, retrieve various sample sets for analysis.

Further advancement of the VepKar corpus will presumably proceed in two directions: volume and structure.

Filling of the corpus with new tagged texts i.e., volume augmentation, is limited only by the physical capacity of experts, since proofreading, editing and homonymy elimination are done manually by a small group of linguists. Certain limitations are imposed also by copyright arrangements.

Project participants are dealing with a range of tasks related to expanding the functionality of the corpus. One of the plans e.g., is to add an audio recording and playback module, permitting linguists to record and users to listen to real Karelian and Veps speech. The syntactic markup module, which works with the output of morphological analysis, can be used to study the syntactic links between lexical units. Semantic markup, in turn, can expand the range of studies based on the corpus data to cover other areas in the humanities (literature studies, folklore studies, ethnology, history, etc.).

The tools already available in the VepKar corpus offer unique opportunities for handling quite a number of linguistic tasks in the study of Karelian and Veps vocabulary and grammar. Material from the corpus is used in the making of new dictionaries and grammar books, and in the process of editing the rules and norms of newly written varieties of the languages in question. The corpus can also be a source of material for Karelian and Veps language instruction. Besides, data from the VepKar platform, provided that their volume and structural potential are further expanded, can be of use in dealing with complicated linguistic tasks such as the development of automatic tools for spellcheck, speech recognition and synthesis, machine translation, building of computer models of the Karelian and Veps languages.

**Funding**

The study is carried out with regular budget funding to the Karelian Research Centre RAS.